\documentclass[sigconf]{acmart}

\AtBeginDocument{%
  \providecommand\BibTeX{{%
    \normalfont B\kern-0.5em{\scshape i\kern-0.25em b}\kern-0.8em\TeX}}}

\usepackage{color, colortbl}
\definecolor{Gray}{gray}{0.8}

\copyrightyear{2022} 
\acmYear{2022} 
\setcopyright{acmcopyright}\acmConference[FDG '22]{FDG '22: Proceedings of the 17th International Conference on the Foundations of Digital Games}{September 5--8, 2022}{Athens, Greece}
\acmBooktitle{FDG '22: Proceedings of the 17th International Conference on the Foundations of Digital Games (FDG '22), September 5--8, 2022, Athens, Greece}
\acmPrice{15.00}
\acmDOI{10.1145/3555858.3555879}
\acmISBN{978-1-4503-9795-7/22/09}

\begin{document}

\title{Generative Personas That Behave and Experience Like Humans}

\author{Matthew Barthet}
\affiliation{%
  \institution{Institute of Digital Games}
  \streetaddress{University of Malta}
  \city{Msida}
  \country{Malta}}
\email{matthew.barthet@um.edu.mt}

\author{Ahmed Khalifa}
\affiliation{%
  \institution{Institute of Digital Games}
  \streetaddress{University of Malta}
  \city{Msida}
  \country{Malta}}
\email{ahmed.khalifa@um.edu.mt}

\author{Antonios Liapis}
\affiliation{%
  \institution{Institute of Digital Games}
  \streetaddress{University of Malta}
  \city{Msida}
  \country{Malta}}
\email{antonios.liapis@um.edu.mt}

\author{Georgios N. Yannakakis}
\affiliation{%
  \institution{Institute of Digital Games}
  \streetaddress{University of Malta}
  \city{Msida}
  \country{Malta}}
\email{georgios.yannakakis@um.edu.mt}

\renewcommand{\shortauthors}{Barthet, et al.}

\begin{abstract}

Using artificial intelligence (AI) to automatically test a game remains a critical challenge for the development of richer and more complex game worlds and for the advancement of AI at large. One of the most promising methods for achieving that long-standing goal is the use of generative AI agents, namely \emph{procedural personas}, that attempt to imitate particular playing behaviors which are represented as rules, rewards, or human demonstrations. All research efforts for building those generative agents, however, have focused solely on playing \emph{behavior} which is arguably a narrow perspective of what a player actually does in a game. Motivated by this gap in the existing state of the art, in this paper we extend the notion of behavioral procedural personas to cater for player experience, thus examining generative agents that can both behave and experience their game as humans would. For that purpose, we employ the Go-Explore reinforcement learning paradigm for training human-like procedural personas, and we test our method on behavior and experience demonstrations of more than 100 players of a racing game. Our findings suggest that the generated agents exhibit distinctive play styles and experience responses of the human personas they were designed to imitate. Importantly, it also appears that experience, which is tied to playing behavior, can be a highly informative driver for better behavioral exploration. 

\end{abstract}




\maketitle

\begin{figure}[h]
\centerline{\includegraphics[width=1.0\columnwidth]{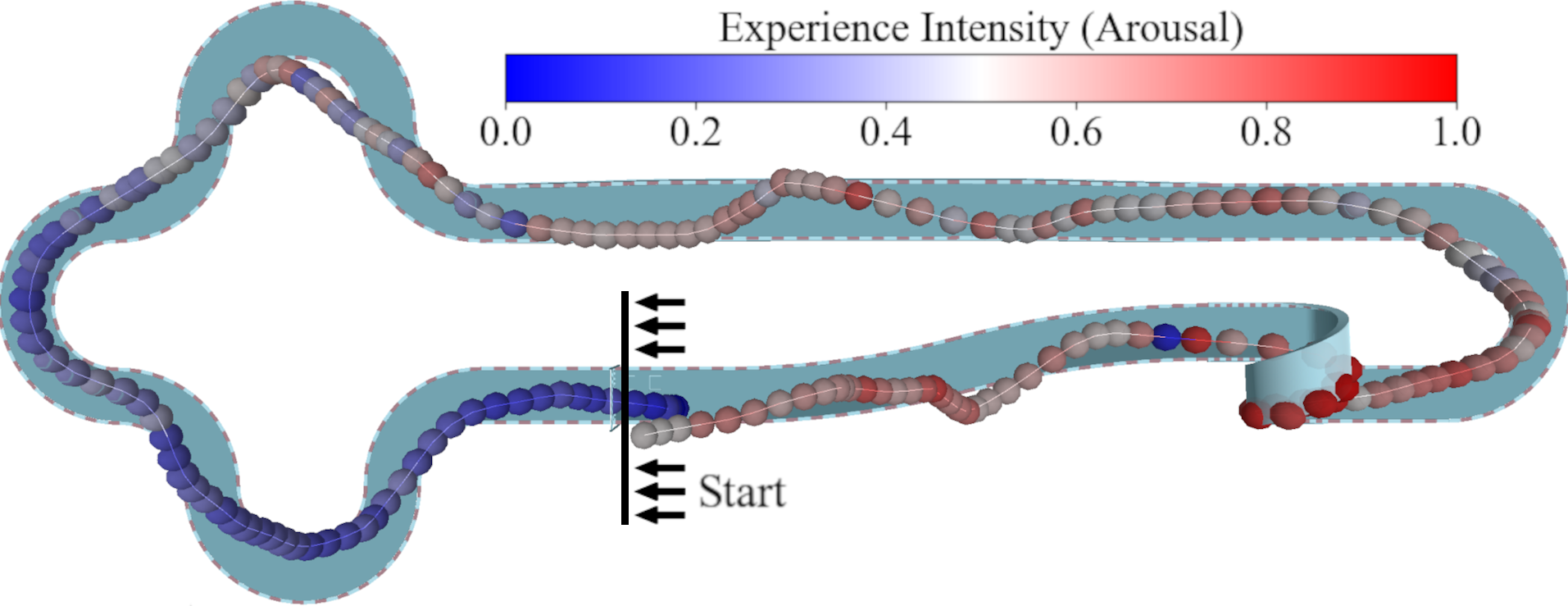}}
\caption{A generative car racing persona that imitates the behavior and experience of a human player using the \textit{Go-Blend} method proposed in this paper. Behavior is depicted as a trace of waypoints and experience is illustrated as color: blue and red waypoints correspond, respectively, to low and high emotional intensity (arousal) during the game.}
\label{fig:solid_trace}
\end{figure}

\section{Introduction}

Players' behavioral and emotional reactions to in-game stimuli often vary significantly between individuals. Professional players will likely experience and react to in-game events far differently than newcomers in most games, as the game becomes less challenging and their play-style becomes more complex or creative. During game production, game designers theorize about the intended play-styles and design the game to align with these theoretical styles, in a form of top-down persona design \cite{canossa2009patterns}. When the game is released, one may observe the playtraces of different players and attempt to cluster them as different behavioral play-styles. These styles might match the intended ones; however, it is expected that entirely new play-styles will emerge. These different styles are often referred to as \emph{player personas} \cite{canossa2009patterns}.

\emph{Procedural personas} are generative AI agents that are able to match the behavior of \emph{player personas} defined in a top-down fashion (from designer intents)~\cite{liapis2015procedural} or a bottom-up fashion (from player data) \cite{holmgaard2014evolving,holmgaard2018automated}. These agents, however, have so far been designed based solely on \emph{behavioral} aspects of playing a game (i.e. what a player does), thereby, ignoring largely the \emph{experience} of play (i.e. how a player feels) \cite{yannakakis2018artificial}. We argue that player experience demonstrations---in addition to behavioral demonstrations---could only enhance the expressive capacity and testing ability of generative personas. Such agents would be able to test game content traditionally by \emph{behaving} as humans would, but importantly also \emph{experience} the game as human players would (i.e. in ways that human experience demonstrations suggest). 

Motivated by the lack of studies for generative personas that both behave and experience their worlds, in this paper we use Go-Explore \cite{Go-Explore}---a recent cutting-edge reinforcement learning algorithm---to build agents who imitate the behavior and experience of human personas, and we test the algorithm on an arcade racing game. The game is accompanied by a dataset \cite{AGAIN} of over 100 human playtraces, containing gameplay data and annotations of arousal (i.e. emotional intensity) provided by the players themselves. We first identify human personas through a data-driven approach, by aggregating the human session data and performing agglomerative hierarchical clustering. We then train agents to mimic the behavior and experience of the identified personas of the game. 

This work builds upon and extends significantly the work of Barthet et al. \cite{barthet2021go} by using Go-Explore for imitating humans across both behavior and experience, namely \emph{Go-Blend}, in a more complex, fast-paced, continuous-control game. The results from our case study show that Go-Blend is capable of generating trajectories which exhibit significantly different behaviors and experiences based on the persona-specific reward function used during exploration. Surprisingly, it seems that for some human player personas, experience (as a reward function) can be the driver for agents that play the game better, thereby, offering insights on the unknown relationship between behavior and experience in play. By extending these tests into stochastic settings, and by using more complex representations of player behavior, we argue that the proposed framework can be a powerful tool for human-like automated play-testing and experience-driven content generation \cite{yannakakis2011experience,shu2021experience}.

\section{Background}

This section provides a brief overview of the use of Reinforcement Learning in games with a focus on the Go-Explore algorithm (Section \ref{sec:RL}), and reviews the various uses of player (and persona) experience modeling in games (Section \ref{sec:personas}).

\subsection{Reinforcement Learning and Go-Explore} \label{sec:RL}

Reinforcement learning (RL) is a popular family of machine learning algorithms which lean on the perspective of behavioral psychology. RL agents typically learn a policy through trial and error, receiving positive or negative rewards for their actions~\cite{RL}. RL has traditionally been used in games for {optimal play} (i.e. playing to win), where the agent learns to play the game as efficiently as possible \cite{yannakakis2018artificial}. Notable achievements in recent deep RL algorithms include super human levels of performance in games such as Go \cite{RL-GO}, Atari Games~\cite{mnih2013playing}, Dota II \cite{RL-Dota}, and Starcraft \cite{RL-Starcraft}. Beyond learning to play games optimally, RL has also been used to imitate human behavior~\cite{Backward,holmgaard2014evolving}, for generating content~\cite{khalifa2020pcgrl,shu2021experience}, and as the underlying method for mixed-initiative design tools~\cite{delarosa2021mixed,guzdial2019friend}. Whilst the use of human-annotated emotions as a training signal remains limited \cite{moerland2018emotion}, the use of simulated affect signals in training has been demonstrated for social referencing in simple robotics tasks \cite{hasson2011emotions} as well as to help accelerate training and avoid premature convergence \cite{broekens2007affect}.

RL algorithms usually struggle in hard-exploration tasks containing {sparse} and {deceptive} rewards \cite{anderson2018deceptive}, and may suffer from \emph{derailment} and \emph{detachment} during learning. Detachment occurs when there are multiples areas of the search space to explore, resulting in agents that forget how to reach previously explored areas. Derailment occurs when distant states are very difficult to reach during training due to a high probability of exploratory actions preventing it from being reached. Algorithms such as Go-Explore \cite{Go-Explore} and \textit{BeBold} \cite{Bebold} are recent RL frameworks specifically designed to tackle these issues. 

In this paper, we focus on the exploration phase from Go-Explore, which builds an archive of promising game states by exploring the environment and storing the states which have the best reward. At its most basic form, exploration is done by randomly selecting a state from the archive, returning to it by replaying its trajectory and exploring a fixed number of random actions before selecting a new state (see section \ref{sec:algorithm}). In Go-Explore, a trajectory refers to the sequence of states and actions required to be taken by an agent to reach a given state in a deterministic setting.

Go-Explore has been demonstrated to achieve previously unmatched performance in challenging Atari games such as \emph{Pitfall} and \emph{Montezuma's Revenge}, which feature sparse deceptive rewards that, in turn, result in premature convergence of deep RL algorithms. The performance of Go-Explore has been demonstrated in text-based games where it outperformed traditional agents in \emph{Zork I} \cite{ammanabrolu2020avoid}, and has shown its ability to generalize to unseen text-based games more effectively \cite{madotto2020exploration}. The algorithm's capabilities have also been demonstrated in complex maze navigation tasks which could not be completed by traditional RL agents \cite{matheron2020pbcs}. Beyond playing planning-based games with exceptional performance, Go-Explore has also been used for autonomous vehicle control for adaptive stress testing \cite{koren2020adaptive}, and as a mixed-initiative tool for quality-assurance testing using automated exploration \cite{chang2019reveal}.

Barthet et al.~\cite{barthet2021go} introduced Go-Blend, a proof-of-concept study that used Go-Explore to model affect as an RL process. The outcomes of that study were trajectories that were able to blend behavioral rewards (i.e. trajectories that play optimally) with affective rewards (i.e. trajectories that ``feel'' like a human would). In this work, we build upon and extend significantly the Go-Blend framework to a real-time and complex game that features a continuous action space. Moreover, we attempt to generate trajectories that both behave and experience a game as humans would, based on a dataset of human demonstrations and annotations.

\subsection{Personas and Player Experience} \label{sec:personas}

Human-computer interaction research has identified \emph{personas} as a grouping of individuals based on a set of descriptors~\cite{cooper2007face}; this can be done in the design phase by crafting a synthetic persona, or as a categorization of a market segment of an already launched product. A \emph{player persona} is the extension of the persona notion in the context of games and refers to categorization methods of players based on their interaction with the game~\cite{canossa2009patterns}. Traditionally, personas can be created using one of two high-level approaches: a \textit{model-based} (top-down) design or a \textit{model-free} (bottom-up) design \cite{yannakakis2013player,yannakakis2018artificial}. Holmgard et al. introduced the notion of \emph{procedural personas}; generative AI agents that learn to play according to different designer goals \cite{holmgaard2014evolving} or human playtraces \cite{holmgard2015evolving}. Procedural personas can be used for automated play testing with different play styles, and for evaluating content through Experience-Driven Procedural Content Generation (EDPCG) algorithms \cite{yannakakis2011experience}. 

Modeling players' experience in games is an active research topic that is also tied heavily to automatic play testing~\cite{holmgaard2018automated,perez2021generating,holmgaard2014evolving} and EDPCG~\cite{liapis2015procedural,fernandes2021adapting}. Early work in the field demonstrated how machine learning models could be trained to predict player frustration, challenge and fun, which can, in turn, be used as fitness functions for generating levels in games such as \textit{Super Mario Bros} (Nintendo, 1985) \cite{pedersen2009modeling}. Beyond its many applications of automatic level generation, EDPCG has also been applied to mixed-initiative level design \cite{yannakakis2014mixed, liapis2013sentient} and domains such as generative music \cite{plans2012experience} and visuals \cite{liapis2012adapting}. Experience-Driven Procedural Content Generation via Reinforcement Learning (EDRL) \cite{shu2021experience} further expands upon this framework by fusing EDPCG~\cite{yannakakis2011experience} and procedural content generation via reinforcement learning \cite{khalifa2020pcgrl}. In the initial EDRL study by Shu et al. \cite{shu2021experience}, agents were trained via RL to design personalized levels for \textit{Super Mario Bros} in an online fashion by maximizing a quantified notion of ``fun'' as described by Koster \cite{koster2013theory}.

In this paper, we focus on data-driven, bottom-up approaches to procedural persona modeling, specifically through clustering of human demonstrations of behavior~\cite{zhang2016data}. We create trajectories that both behave and ``experience'' like human players (personas), thus generating trajectories with a diverse set of play-styles and affective response patterns. Whilst we do not tackle any form of content generation in this paper, our Go-Blend trajectories can be used to train agents for automatic level testing and content evaluation.

\begin{figure}[!tb]
\centering
\includegraphics[width=.7\columnwidth]{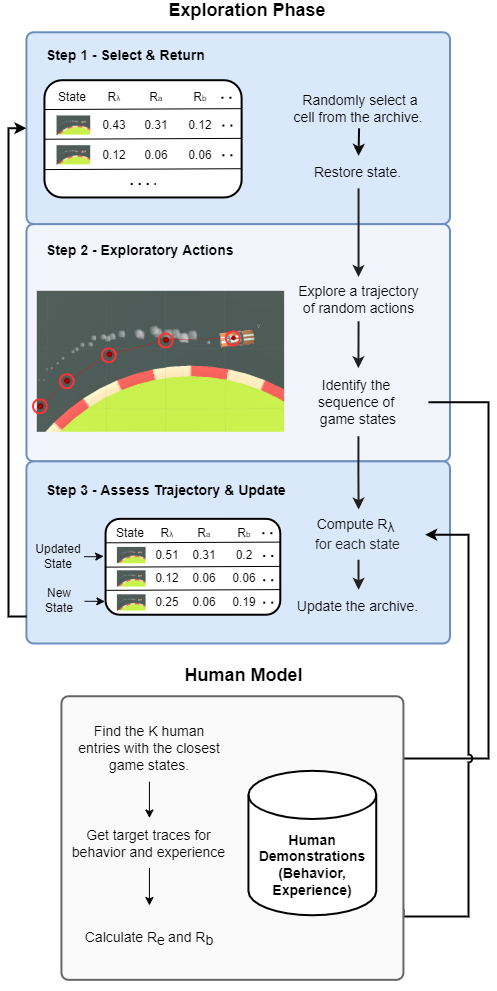}
\caption{A high-level overview of Go-Blend for human imitation. The framework generates trajectories that imitate human-like behavior and experience.}
\label{fig:algorithm}
\end{figure}

\section{Go-Blend for Procedural Personas}

In this paper, we expand upon the Go-Blend framework~\cite{barthet2021go} by imitating different human personas across both behavioral and experience dimensions in a challenging racing game. In this section, we go through the basic steps of the Go-Blend algorithm (Section~\ref{sec:algorithm}), we outline the reward signals used  (Section~\ref{sec:reward}), and the ways those rewards are integrated in the algorithm. Figure~\ref{fig:algorithm} illustrates the core aspects of the Go-Blend framework. 

\subsection{Algorithm}\label{sec:algorithm}

Go-Blend~\cite{barthet2021go} is an implementation of the Go-Explore algorithm that not only builds trajectories that behave in certain ways but also optimizes aspects of their player experience. The implementation of Go-Blend follows closely the original implementation of Go-Explore proposed by Ecoffet et al.~\cite{Go-Explore}. Due to the deterministic nature of the test-bed game of this paper, we focus primarily on the exploration phase of the algorithm. During exploration, the system maintains an archive of cells, each containing a unique game state that has been observed so far. Each cell also contains the trajectory of actions required to return to its game state, behavior reward (e.g. game score), and experience reward (e.g. annotated arousal). These cells are selected using a desired cell selection strategy (e.g. random, tournament selection, UCB), after which the system returns to the chosen state and begins exploring from there. Exploratory actions also follow their own action selection mechanism (e.g. random, domain knowledge, policy-based) and are used to expand upon the current cell's trajectory.

After each exploratory action, a cell is constructed according to the current game state. Each cell has an associated reward value ($R_{\lambda}$) which is used to determine if the archive should be updated. Cells are always added to the archive if they have not been encountered so far. If the cell exists in the archive, it is updated only if it satisfies one of two replacement criteria: a) any cell encountered with a better $R_{\lambda}$ than its existing counterpart in the archive is updated, b) any cell with the same $R_{\lambda}$ as the existing cell is updated if it has a more efficient (shorter) trajectory. As Go-Blend considers both play behavior and experience, $R_{\lambda}$ is calculated using corresponding reward functions $R_{b}$ and $R_{e}$ (see section \ref{sec:reward}) saved in each cell. These values can be used for selection and replacement to prioritize cells with the desired qualities. 

The process of selecting cells, returning to their state and exploring new actions is repeated for a fixed number of iterations or until a desired stopping criterion is reached (e.g. reaching the optimal score). The result of this exploration phase is a number of high-performing cells for the given deterministic environment. If required, these trajectories can be used to create a robustified agent using imitation learning or RL algorithms such as the ``Backwards Algorithm'' \cite{Backward}. This optional step creates an RL agent that is capable of performing at the level achieved during exploration in a stochastic environment. As mentioned, this step is not necessary in the deterministic game test-bed of this paper, and is not implemented.

\subsection{Reward Functions}\label{sec:reward}

As mentioned before, we extend Go-Blend to not only reward imitating human traces of experience ($R_e$), but also imitating traces of their behavior ($R_b$). Therefore, both reward functions are calculated using the same formula (see Equation~\ref{eq:imitation_behavior}), which we call $R_x$. In short, Eq.~\eqref{eq:imitation_behavior} calculates the average similarity between the data points in the target (human) trajectory, and the data points in the Go-Blend trajectory.

\begin{equation}
R_{x} = \frac{1}{n} \sum^{n}_{i=0} \left(1 - |h_{x}(i)-t_{x}(i)|\right)^2
\label{eq:imitation_behavior}
\end{equation}
\noindent where $n$ is the number of observations (time windows) made so far in this trajectory; $i$ is the time window being evaluated; $h_{e}(i)$ is the  experience metric for time window $i$ and $h_{b}(i)$ is its behavior metric; $t_{e}(i)$ and $t_{b}(i)$ is the target experience value and target behavior value respectively for time window $i$. These target values are derived from the human model, as the two rewards aim to minimize discrepancy between the closest human behavior of experience per time window that has occurred so far within the trajectory.

Our implementation combines the $R_b$ and $R_e$ components into a single, weighted reward function as seen in Go-Blend~\cite{barthet2021go}. Both reward components are normalized within the range $[0,1]$ to avoid uneven weighting between the two objectives. The reward function, denoted by $R_\lambda$, is formally defined as follows:
\begin{equation}
    R_\lambda = \lambda \cdot R_e + (1-\lambda) \cdot R_b
    \label{eq:lambda}
\end{equation}
\noindent where $\lambda$ is the weight parameter that blends the two components; $R_e$ and $R_b$ correspond to the experience and behavior reward, respectively. By increasing $\lambda$, we instruct Go-Blend to increasingly prioritize imitating player experience, and vice versa. At $\lambda=0$, the trajectories are rewarded solely on imitating human behavior, whilst at $\lambda=1$ they are rewarded on just their experience imitation.

Both $R_b$ and $R_e$ therefore assume time-continuous signals for either behavior or experience respectively. The AI agent produces a time-continuous signal, which should match human-provided time-continuous signals (e.g. from a single human player that the agent imitates, or aggregated from many players). The distance is squared to penalize larger deviations from the target signal even more during exploration, whilst making it easier to visually distinguish between good and bad performing trajectories during evaluation. Since both $R_e$ and $R_b$ calculate the average distance across all the observations made so far, it encourages trajectories with high imitation accuracy across the entire duration of the game. 

As mentioned, $R_e$ is the reward for imitating human experience. This function can be used for imitating any form of measure for experience, such as frustration, arousal, and fun, given the appropriate model for mapping game states to player experience (see Figure \ref{fig:algorithm}). Similarly, as $R_b$ is the reward for imitating human behavior, any measure for behavior can be used such as player inputs, score traces, and game mechanics' frequencies. For more complex behavior or experience imitation, $h_{x}(i)$ and $t_{x}(i)$ can be seen as a vector for all the metrics that should be imitated, in which case equation~\eqref{eq:imitation_behavior} calculates the vector distance.

\section{Test-Bed Racing Game: Solid Rally} \label{sec:solid}

We test our proposed system for imitating behavior and emotion of player clusters in the ``Solid Rally'' driving game (hereafter \emph{Solid}). Solid is a racing game built using the Unity Engine and forms part of the nine games featured in the AGAIN dataset \cite{AGAIN}. This game was chosen for its non-trivial mechanics and objective (beat the opponent cars within the time limit), as well as its associated dataset of 108 human demonstrations with annotated arousal traces. In this section, we describe the properties of the game and its human playtraces, how the game-states are represented for Go-Blend along with the rewards used, and finally how we split the human playtraces into player personas in order to derive persona-specific rewards for the Go-Blend algorithm. 

\begin{figure}[!tb]
\centering
\includegraphics[width=0.85\columnwidth]{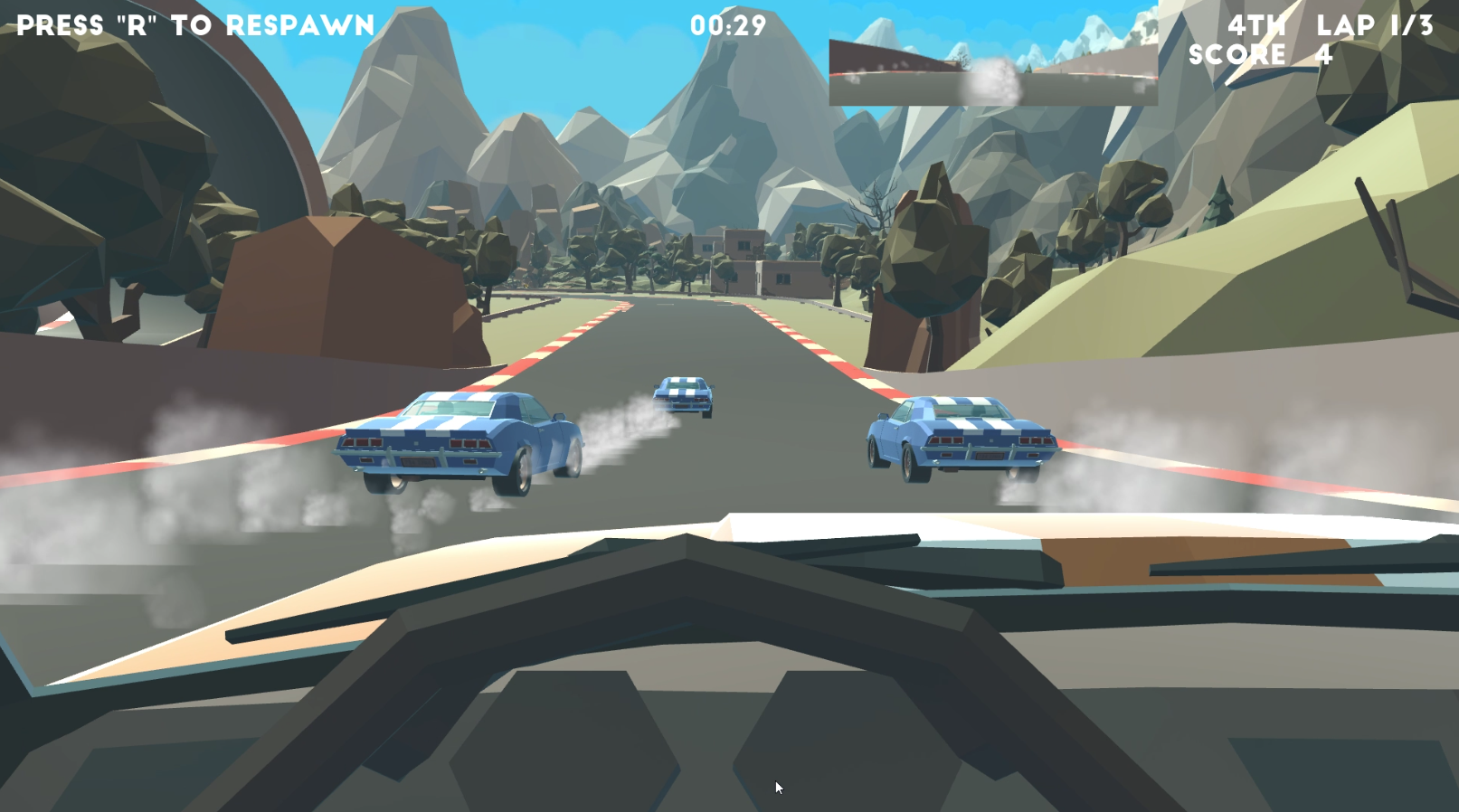}
\caption{First-person view in the Solid Rally racing game.}
\label{fig:solid}
\end{figure}

\subsection{Game Description}

In Solid, the player controls a rally car from a first-person perspective (as shown in Figure~\ref{fig:solid}) and attempt to end the race ahead of the three opponent cars. The race ends when the player has completed three laps or has driven for two minutes, after which they are given a placement and their points tally into a \textit{final score}. Points are awarded for successfully driving around the circuit and passing through checkpoints. There are 8 checkpoints per lap which results to a maximum score of 24 points awarded for completing the full three laps within the time limit. The player controls the car's gas pedal and steering wheel through the arrow keys. There are three possible inputs for steering (-1, 0, 1) and gas (-1, 0, 1), where 0 is the neutral state when no key is pressed. The car's handling loosely simulates a rally car, meaning the player must balance the gas and steering input to drift the car past corners. The track contains ``off-track'' grass segments which slow the cars down slightly if driven over, providing a viable strategy to cut corners in the circuit. The AI opponent cars use a simple deterministic controller which follows a set of waypoints to drive around the circuit.

This game is accompanied by a dataset of 108 human play sessions (excluding outliers) with player annotated arousal traces, called the AGAIN dataset \cite{AGAIN}. Each entry in the dataset contains the average data for the previous 250 milliseconds across 32 in-game features specific to Solid. These features cover basic spatial properties such as rotation, speed, and collision status for both the player and opponent cars, as well as their current position, score, steering and pedal input. Arousal traces were collected in a continuous, unbounded fashion using RankTrace \cite{RankTrace} and the PAGAN \cite{PAGAN} online annotation framework. These arousal traces were then normalized on a per-session basis to account for discrepancies in the value ranges between players.

Note that while the playtraces and arousal annotations in AGAIN cover games that may last three laps or up to 2 minutes, in reality many of the players did not complete all three laps within the time limit. For Go-Blend we therefore use the behavior and experience data of the first two laps, and test the performance of the AI personas on races that last only two laps in this paper.

\subsection{Go-Blend for Solid Rally}\label{sec:go-solid}

To use Go-Blend, the game must be represented in a way that can be mapped to a cell in the archive of game states. The in-game features, provided by AGAIN, however, lack the granularity to adequately distinguish between meaningfully different game states. Therefore, a new state representation was defined using categorical variables. The player car's \textit{speed} is distinguished between two states (slow, fast) and its \textit{rotation} between six 30-degree thresholds. The circuit is split into 19 segments according to their high-level structure (e.g. straight, half-curve, full-curve) which are further split into sub-segments according to their shape (e.g. left side and right side, off-road) to identify the player's location on the track. Finally, cells are also distinguished by the player's lap number and their proximity to opponent cars (i.e. proximity is true if an opponent car is on the same sub-segment). This results in $4,800$ possible cells in the archive for a race distance of two laps.

Due to the game's reliance on the physics system of Unity, some changes had to be made to the settings of the game engine and car controller scripts to maximize the determinism of the environment. The game was set to a ``forced'' frame-rate mode to ensure the same number of frames occur between each event. The game engine was set to ``enhanced determinism'' mode with the physics precision increased. These changes were important to ensure that the trajectories found through Go-Explore's exploration phase were replayable without any significant deviation from the results seen during exploration. We use the player's score as our measure of in-game behavior where the goal is to reach as many checkpoints as possible. Specifically, we compute $h_{x}(i)$ in equation~\eqref{eq:imitation_behavior} as the number of checkpoints crossed so far in the current trajectory until time window $i$, divided by the maximum possible score for two laps (16 points) in order to derive a value normalized within $[0,1]$.

As mentioned, we use the annotated arousal traces provided with the human demonstrations for Solid as our measure of player experience. Our approach for calculating the current arousal value for the player, i.e. $h_{x}(i)$ in equation~\eqref{eq:imitation_behavior}, is similar to that used by Barthet et al.~\cite{barthet2021go}. The AGAIN dataset \cite{AGAIN} provides human experience demonstrations in the form of moment-to-moment (i.e. 4Hz) players' annotated arousal values which are linked to a vector of 32 in-game features. We directly use these arousal labels to build our arousal reward functions rather than relying on a trained surrogate model of arousal to predict outcomes indirectly. 

After each exploratory action taken the algorithm queries the Solid dataset for the arousal values of the K-nearest neighbors. The current arousal at this time window $i$, i.e. $h_{e}(i)$ in equation~\eqref{eq:imitation_behavior}, is taken as the mean of these arousal values, which is calculated using distance-weighted kNN \cite{dudani1976distance}. The $k$ nearest neighbors are found by calculating the Euclidean distance from the current feature vector to every entry in the dataset of Solid playtraces. To find the weighted average, each arousal value is weighted by their corresponding distance to give more weight to the arousal values of the closest neighbors. These weighted values are added together and divided by the sum of the neighbor distances to obtain a value normalized within $[0,1]$. 

In short, this method of calculating the arousal reward for Go-Blend reduces the noise caused by outliers in the AGAIN playtraces, and biases arousal similarity towards those playtraces that are most similar to the agent's game-state during each time-window. Note that the target arousal and behavior per time window ($t_e(i)$ and $t_b(i)$ respectively) are calculated based on the personas discovered from the AGAIN playtraces as described in section \ref{sec:clustering}.

\begin{figure}[t]
\centerline{\includegraphics[width=0.9\columnwidth]{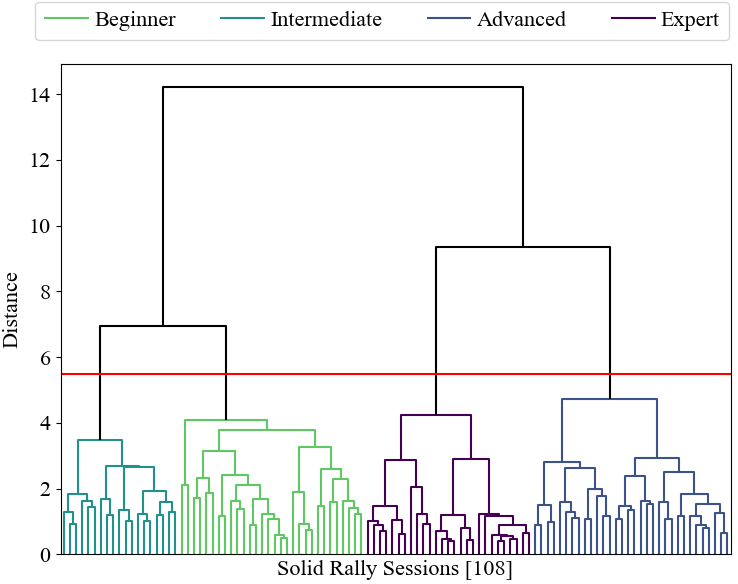}}
\caption{Clustering of aggregated session data using the ward dendogram method and a T-value of 5.5 (red horizontal line) revealing 4 clusters.}
\label{fig:dendogram}
\vspace{-10pt}
\end{figure}

\subsection{Player Personas in Solid Rally}\label{sec:clustering}

\begin{figure}[t]
\centerline{\includegraphics[width=0.9\columnwidth]{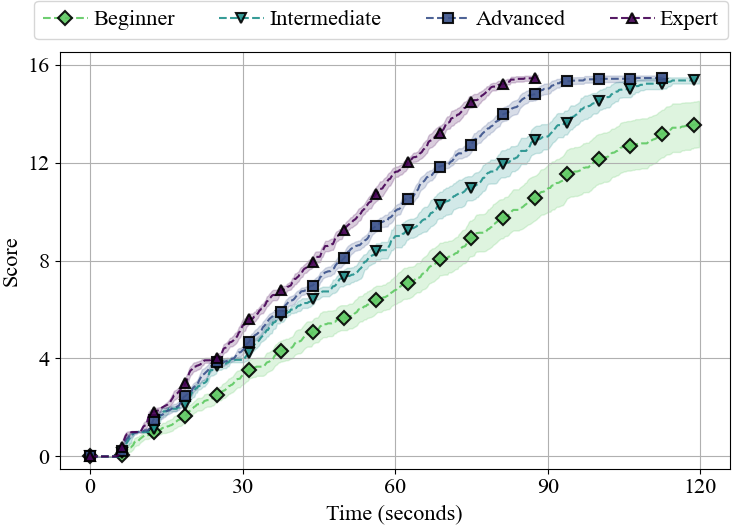}}
\centerline{\includegraphics[width=0.91\columnwidth]{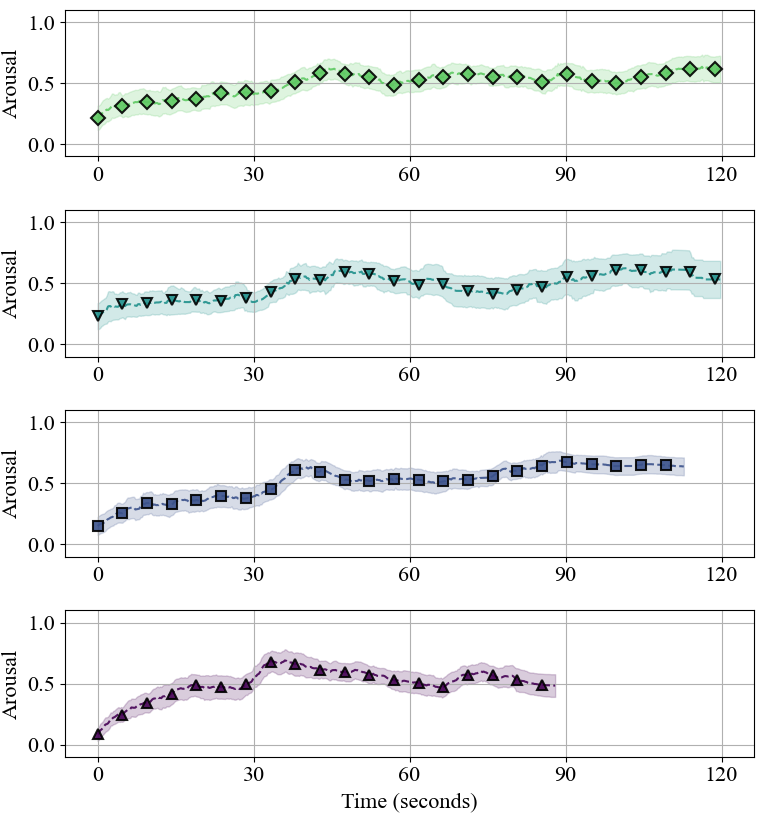}}
\caption{Mean in-game score and arousal traces over time for the four identified player personas. Note that a shorter score trace indicates that the race distance has been completed before the time limit by the entire cluster.}
\label{fig:cluster_scores}
\vspace{-10pt}
\end{figure}

In this section, we describe our data-driven approach to identifying player personas from Solid's dataset of 108 human play sessions. Since each play session is made up of a maximum of 480 (250ms) time windows, they need to be aggregated into a single vector that represents the entire session. First, all the play sessions in the dataset were truncated to only include the data for two laps to line up with our experimental protocol. The aggregation methods we used for each feature in the dataset varied based on their type. In the case of the player's score, we take the maximum score achieved in the session, which always corresponds to the score observed in the final time window. Scalar variables such as player speed, distance and rotation changes were represented by their mean. Categorical variables such as the off-road and collision flags were converted into integer numbers representing the frequency of their occurrence.

Figure \ref{fig:dendogram} depicts the resulting dendogram when clustering the aggregated AGAIN dataset of 108 playtraces (i.e. behaviors) using Ward's hierarchical clustering method \cite{Ward} and the squared Euclidean distance as a measure of dissimilarity between feature vectors \cite{Tomb-Raider}; the obtained dendogram shows four distinct behavioral clusters. %We used this number to perform agglomerative hierarchical clustering on the aggregated data.
Based on the four different clusters of players identified, we aim to identify the differences between clusters and provide them with some human-readable labels. Such labels are helpful when defining play personas from data \cite{canossa2009patterns}. Figure~\ref{fig:cluster_scores} shows the average performance (game score) per cluster across a two-lap race in Solid. Based on their performance, we labeled the four clusters as ``expert'' (27 players), ``advanced'' (32 players), ``intermediate'' (19 players), and ``beginner'' (30 players). All player clusters except for the ``beginner'' succeed in completing two laps before the time limit, but some members of the ``beginner'' cluster do not finish the race on time (i.e. the average score is below the maximum of $16$). Beyond the scores across the four clusters, Figure~\ref{fig:cluster_scores} also displays the corresponding \emph{mean arousal traces} for each cluster. This figure indicates that there is no apparent linear correlation between Solid's feature set and the arousal values generated. In all experiments reported in this paper, the mean score trace and mean arousal trace of each persona (i.e. cluster of players) are used as target values ($t_{e}(i)$ and $t_b(i)$) for behavior and arousal persona imitation, respectively. 

\section{Experiments} \label{sec:exp_port}

In this paper, we introduce the notion of a Go-Blend RL algorithm that is guided by a player persona rather than the entire player population. We expect that a player persona can provide a more coherent playtrace for behavioral imitation, and that similarly the experience of the persona's cluster will also be idiosyncratic to the experience of the players within the same (behavioral) cluster. Our core dimension of inquiry is whether Go-Blend agents that aim to match either the behavior or the experience or a combination thereof will differ depending on which persona they are imitating. A complementary inquiry is whether some personas are better at providing guidance in terms of in-game performance, since in Fig.~\ref{fig:cluster_scores} we established that the ``expert'' persona cluster has better in-game scores than others. In our experiments, we varied $\lambda$ in Eq.~\eqref{eq:lambda} within $[0, 0.5, 1]$ to test behavior imitation, blended behavior and arousal imitation, and arousal imitation respectively.

We conducted three independent runs of each experiment for $500,000$ iterations, saving the best cell at the end. During an iteration of exploration, 20 exploratory actions are taken before randomly selecting a new cell to load and explore from. Actions are selected every 250ms using a weighted random selection from the possible inputs for steering (-1, 0, 1) and gas (-1, 0, 1), where the weights are the normalized frequency of the inputs in all 108 human demonstrations in AGAIN. 

For the experience reward $R_e$, we derive the arousal trace of the Go-Blend agent so far through the kNN method described in Section \ref{sec:go-solid}, considering the $k=5$ nearest neighbors from the play sessions belonging to that persona only (i.e. ignoring game states belonging to other personas). For the behavior reward $R_b$ we derive the game score trace normalized to the max score of the trajectory so far, and reward the agent according to how closely it matches with the mean game score of the relevant persona cluster (Figure~\ref{fig:cluster_scores}). 

Beyond comparing between Go-Blend variants trained on different personas and the personas themselves, we also compare against two baseline agents: a \textit{random agent} that chooses a gas/steering action via weighted randomness based on the frequency of all human demonstrations (similar to the exploration of Go-Blend) and a \emph{winner} agent that uses Go-Explore to maximize its score within the 2-minute time limit.

\subsection{In-Game Performance Metrics}

\begin{table*}[!tb]
\caption{Results for every experiment in Solid Rally, grouped by the persona being imitated by each experiment and averaged across 3 runs, including the 95\% confidence interval. Cells colored in gray represent the statistics of the human personas. Bold cells point to in-game statistic values of Go-Blend that are closer to those of the corresponding human persona.
} %whereas bold cells represent the Go-Blend experiments with the result most similar to their persona.}

% \vspace{5pt}
\begin{center} 
\begin{tabular}{|c|c|c|c|c|c|c|c|}
\hline
\textbf{Experiment}&\multicolumn{7}{|c|}{\textbf{In-Game Statistics}} \\
\cline{2-8} 
\textbf{Setup} & Final Score & Lap 1 Time (s) & Average Speed & Nearest Car & Off-Road (\%) & Midair (\%) & Crashing (\%)\\
\hline
\hline
Random & 0 $\pm$0.00 & N/A & 0.98 $\pm$0.01 & 379.93 $\pm$19.75 & 97.78 $\pm$1.06 & 0.21 $\pm$0.0 & 13.61 $\pm$1.66 \\
\hline
Winner & 16 $\pm$0.00 & 58.67 $\pm$5.73 & 33.1 $\pm$1.44 & 409.15 $\pm$44.76 & 9.92 $\pm$3.33 & 0.15 $\pm$0.12 & 15.48 $\pm$2.27\\
\hline
\hline
\rowcolor{Gray}
Beginner & 14.00$\pm$0.94 & 70.51 $\pm$6.76 & 33.8 $\pm$2.16 & 419.61 $\pm$2.83 & 24.88 $\pm$2.92 & 4.8 $\pm$2.02 & 4.79 $\pm$0.7\\
\hline
$R_{0.0}$ & 13.33 $\pm$0.53 &\textbf{ 76.0 $\pm$16.0} & \textbf{32.01 $\pm$0.95} & 295.78 $\pm$91.59 & 5.94 $\pm$1.03 & 2.95 $\pm$1.03 & \textbf{14.76 $\pm$2.38}\\
\hline
$R_{0.5}$ & \textbf{14.33 $\pm$0.53} & 64.67 $\pm$1.04 & 29.07 $\pm$0.86 & \textbf{342.4 $\pm$9.49} & 5.92 $\pm$2.17 & \textbf{4.31 $\pm$0.12} & 17.34 $\pm$3.59\\
\hline
$R_{1.0}$ & 4.33 $\pm$0.53 & N/A & 24.6 $\pm$4.37 & 310.88 $\pm$85.79 & \textbf{10.57 $\pm$11.9} & 3.35 $\pm$0.79 & 21.51 $\pm$8.88\\
\hline
\hline
\rowcolor{Gray}
Intermediate & 16.00 $\pm$0.13 & 53.04 $\pm$2.86 & 39.62 $\pm$0.88 & 291.91 $\pm$4.8 & 21.81 $\pm$2.77 & 3.7 $\pm$1.28 & 5.88 $\pm$0.98\\
\hline
$R_{0.0}$ & 14.00 $\pm$0.00 & \textbf{51.75 $\pm$2.23} & 31.3 $\pm$0.29 & \textbf{329.15 $\pm$57.3} & 4.52 $\pm$2.24 & 4.53 $\pm$1.09 & 14.97 $\pm$2.41\\
\hline
$R_{0.5}$ & \textbf{14.67 $\pm$0.53} & 55.75 $\pm$2.05 & \textbf{32.34 $\pm$0.56} & 416.48 $\pm$14.02 & 6.85 $\pm$0.89 & \textbf{3.28 $\pm$0.57} & \textbf{14.69 $\pm$2.12}\\
\hline
$R_{1.0}$ & 2.67 $\pm$4.27 & 98.92 $\pm$11.92 & 19.36 $\pm$3.92 & 376.38 $\pm$29.73 & \textbf{26.41 $\pm$7.89} & 3.2 $\pm$1.75 & 20.31 $\pm$3.1\\
\hline
\hline
\rowcolor{Gray}
Advanced & 16.00 $\pm$0.10 & 47.12 $\pm$0.94 & 43.63 $\pm$0.55 & 146.64 $\pm$3.35 & 13.41 $\pm$1.57 & 2.57 $\pm$0.84 & 7.07 $\pm$0.75 \\
\hline
$R_{0.0}$ & \textbf{13.67 $\pm$1.92} & \textbf{47.42 $\pm$1.49} & \textbf{33.06 $\pm$3.41} & 328.7 $\pm$10.33 & 4.1 $\pm$0.66 & \textbf{4.95 $\pm$1.65} & 15.87 $\pm$7.61\\
\hline
$R_{0.5}$ & 13.67 $\pm$0.53 & 51.42 $\pm$1.79 & 31.48 $\pm$0.77 & \textbf{322.95 $\pm$66.02} & 6.17 $\pm$3.13 & 5.34 $\pm$1.26 & \textbf{15.65 $\pm$2.38}\\
\hline
$R_{1.0}$ & 7.00 $\pm$6.06 & 93.17 $\pm$31.54 & 22.46 $\pm$5.64 & 389.7 $\pm$13.24 & \textbf{10.19 $\pm$10.42} & 7.09 $\pm$3.38 & 19.92 $\pm$3.4\\
\hline
\hline
\rowcolor{Gray}
Expert & 16.00 $\pm$0.10 & 42.01 $\pm$0.84 & 48.25 $\pm$0.67 & 306.66 $\pm$4.6 & 11.06 $\pm$1.7 & 1.52 $\pm$0.24 & 5.83 $\pm$0.57\\
\hline
$R_{0.0}$ & \textbf{15.00 $\pm$0.00} & \textbf{44.42 $\pm$0.67} & \textbf{42.17 $\pm$0.66} & 41.98 $\pm$19.26 & 2.61 $\pm$1.12 & 3.31 $\pm$0.26 & \textbf{13.07 $\pm$1.06}\\
\hline
$R_{0.5}$ & 14.33 $\pm$0.53 & 44.67 $\pm$0.58 & 38.98 $\pm$2.65 & 97.19 $\pm$53.13 & 4.82 $\pm$3.46 & 4.84 $\pm$1.59 & 16.21 $\pm$1.55\\
\hline
$R_{1.0}$ & 11.00 $\pm$2.44 & 48.08 $\pm$2.79 & 33.33 $\pm$0.25 & \textbf{234.25 $\pm$58.41} & \textbf{13.88 $\pm$6.61} & \textbf{2.93 $\pm$1.06} & 15.86 $\pm$0.45\\
\hline
\end{tabular}
\label{tab:performance_metrics}
\end{center}
\vspace{-10pt}
\end{table*} 

Table~\ref{tab:performance_metrics} shows a set of performance metrics for all twelve experiments as they are compared to their respective personas (clusters of human behaviors), and the two AI baselines. The performance metrics include the final score achieved, the agent's average speed during the race, and the time taken to complete the first lap. We also include the agent car's distance to the nearest visible opponent on the track at any given time; if no opponents are visible, this value is set to the maximum possible distance of 500 units. Finally, we include the percentage of time windows spent off-road (i.e. grass segments), midair, and crashing with another opponent or a barrier.

Looking at the table it is immediately apparent that the trajectories generated by Go-Blend differ significantly in behavior based on the persona being imitated. Our $R_{0.0}$ and $R_{0.5}$ experiments produce very similar lap times and average speed around the circuit to their target personas, indicating they are able to scale up and down in performance depending on their target score trace. The main reason for this is the high correlation between these two performance measures and the players' score (e.g: higher score traces need shorter lap times and faster speeds). There is also some variation in play style between the experiments as the higher performing personas pressure Go-Blend to create more efficient paths around the circuit, evident in the decreasing time spent off-road and midair as the performance standard increases.

Looking at the final scores we can observe the ``winner'' agent is the only experiment which perfectly matches the ``intermediate'', ``advanced'' and ``expert'' personas.  The Go-Blend experiments for these personas struggle to finish the second lap and converge prematurely to marginally inferior final scores. As expected the random agent is not capable of driving a representative path around the circuit, failing to reach even the first checkpoint. By looking at the Lap 1 time of the ``winner'' agent, we can see that it achieves this final score at a significantly slower pace than the $R_{0.0}$ and $R_{0.5}$ agents for ``advanced'' and ``expert'' players. This indicates that whilst the ``winner'' agent can very easily explore the track and complete the two laps within the set time limit of two minutes, it is unable to improve on the efficiency of the trajectories past that point. This also sheds light on the lower final scores during imitation, as once a high performing, full-length trajectory is produced, the algorithm struggles to improve upon it further. This limitation is likely caused by the sparsity of the score signal (just 16 points), which leaves the algorithm directionless in-between checkpoints.  Another reason is our relatively simple cell selection and replacement strategy which does not prioritize efficiency in a sufficient manner, and needs to be explored further in future studies.

The table also shows that $R_{1.0}$ experiments, unsurprisingly, fail to imitate their persona across all in-game statistics. The $R_{1.0}$ experiment imitating the ``expert'' persona, however, achieves surprisingly strong results across all our measures despite the fact it does not prioritize behavior imitation during exploration. One reason for this is that the expert human persona completes the race significantly quicker than the other personas, thus pressuring the Go-Blend experiments to produce more efficient trajectories during exploration to cater for the shorter time limit. This pressure for more time-efficient trajectories seems to prevent Go-Blend from getting stuck during exploration, allowing it to explore more high-performing states in Lap 2. This is backed up by the fact that both the ``beginner'' and ``intermediate'' $R_{1.0}$ experiments converge to under 60\% exploration of possible states, whereas their ``advanced'' and ``expert'' counterparts both reach 80\% of possible states. Furthermore, looking at the percentage of states explored in Lap 2, the ``beginner'' and ``intermediate'' $R_{1.0}$ experiments both fail to perform any significant exploration, exploring 0\% and 40\% of possible Lap 2 states, respectively. 

\subsection{Reward Comparison}

\begin{figure}[!tb]
\centerline{\includegraphics[width=.88\columnwidth]{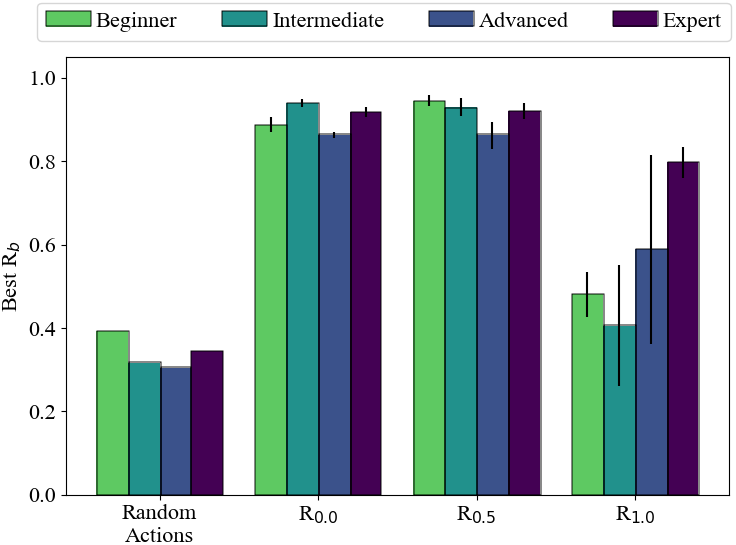}}
\centerline{\includegraphics[width=.88\columnwidth]{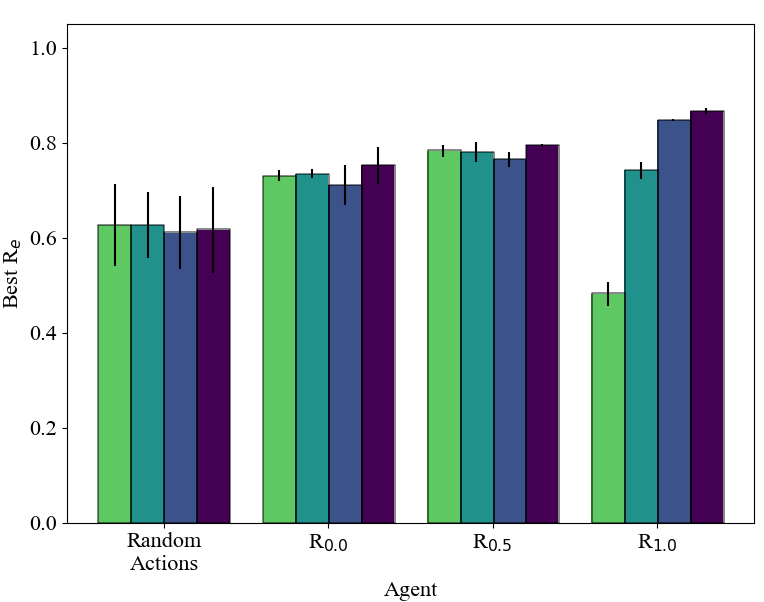}}
\caption{Reward comparison for imitating persona behavior ($R_b$) and experience ($R_{e}$) in Solid, using the best trajectory found in each experiment, averaged across three runs, and including the 95\% confidence interval.}
\label{fig:imitation}
\end{figure}

% \todo{Add black borders to bars in Fig 6}
Beyond the in-game performance of the agents, in this set of experiments we focus on how well our agents match the persona they attempt to imitate in terms of rewards. To assess this, we use the $R_b$ and $R_e$ rewards and evaluate them over the entire trajectory (at the end of the two laps or after the time limit of 2 minutes is elapsed). The reward values of the respective persona each Go-Blend variant aims to emulate are illustrated in Figure~\ref{fig:imitation}, along the reward values of our baseline random agent. We observe that, unsurprisingly, the random agent (even when biased according to the frequency of human actions) does not match the behavior of any persona due to their poor behavior. As corroborated by Table \ref{tab:performance_metrics}, both $R_{0.0}$ and $R_{0.5}$ agents match the performance of the persona they aim to imitate, but the $R_{1.0}$ agents who are trained to imitate only arousal of that persona, unsurprisingly, cannot match its performance. When imitating solely arousal for the ``expert'' persona (i.e. the most proficient players), however, the agent's behavior is surprisingly similar to that of the ``expert'' persona, as seen in the previous section.

In terms of arousal imitation, it is surprising that even the random agent can reach high $R_e$ values across all personas. The most apparent reason for this is that the mean arousal traces for all four personas are relatively stable and hover around the median $(0.5)$ across the duration of the race. Importantly, the random agent employs the same arousal model (i.e. using distance-weighted kNN) as the Go-Blend agents, and thus it seems it is capable of producing a somewhat decent experience trace compared to the target persona, even if it fails to yield a representative behavior. The $R_{0.0}$ agents, however, manage to better match their respective persona's arousal trace compared to the random agent even though they do not consider arousal as a reward. Interestingly, the $R_{1.0}$ experiments imitating the ``beginner" persona produce a much larger discrepancy in arousal (i.e. lower $R_e$) compared to the same experiment for other personas. This is because of this agent's inability to explore past Lap 1 and thereby converging prematurely to a short trajectory which does not even reach the time limit. This results in a poor average score that is significantly worse than any other experiment, including the random agent baseline.

Finally, we observe that arousal imitation is more successful for the most proficient players. Specifically, the $R_{1.0}$ agents trained on the ``expert'' persona, this high match in arousal is accompanied by a fairly close match in terms of performance. The $R_{1.0}$ experiments trained on the ``advanced'' persona also achieve good arousal imitation, but are more inconsistent on their behavior imitation task compared to the ``expert'' imitator. This is also likely due to the shorter traces for ``advanced'' personas and even shorter for ``expert'' personas (see Fig.~\ref{fig:cluster_scores}). 
%This finding aligns well with the ``advanced'' human persona displaying shorter playtraces than the personas, but still significantly longer than the ``expert'' persona. As a result, whilst there is a higher pressure for efficiency during exploration, it is not as significant as that seen in the ``expert'' experiments. 

\section{Discussion}

In this paper, we expanded on the Go-Blend framework to generate RL agents that imitate the behavior and experience of human personas in the real-time racing game \textit{Solid Rally}. This constitutes our initial study on blending notions of behavioral and experience persona modeling using RL. Our approach for identifying personas of human players in a bottom-up fashion through the AGAIN dataset yielded 4 distinct groups of players in terms of their behavior, the majority of which Go-Blend was capable of imitating accurately. The high imitation accuracy of Go-Blend agents for both behavior and experience of human personas highlights the potential of Go-Blend for automated play testing and EDPCG, and opens up a number of interesting avenues for future work. In this section we discuss the limitations of our approach and outline the steps to further improve on the efficiency, robustness and scalability of the method.

Our simple reward function for imitating behavior using the personas' mean score produces trajectories which imitate their level of performance, but not necessarily their actual behavior on track. This is evident in the bigger discrepancy between our generated trajectories and the personas in off-road time, distance to the nearest opponent and time spent crashing into other objects. The in-game behaviors which are easiest to imitate are the ones that correlate with score such as the average speed. A more complex reward function for behavior imitation, which incorporates more dimensions than just score could provide more representative behavior for the given persona. One alternative representation of behavior could be imitating the entire feature trace (32 elements) of human players which should provide a richer representation of their behavior on-track. When it comes to experience imitation, adding further affect dimensions to the existing model such as valence and dominance would create agents with richer and wider affect responses to in-game stimuli. Moreover, extending our reward functions to consider the uncertainty of the values generated from our human demonstrations could help the agents identify more promising states with lower disagreement between playtraces and experience annotations.

Our current approach does not consider reward as part of the cell selection strategy for exploration in Go-Explore. In our implementation, we only use the cell reward for replacement in the archive. Selecting cells according to their $R_e$ (e.g. tournament selection, UCB) and leaving cell replacement to their $R_b$ or raw in-game score could be a more efficient alternative to the blended reward function used in this paper. This direction would also allow for a more in-depth evaluation of how player experience drives action selection during gameplay, and could possibly lead to more efficient and robust training, or even discover novel behaviors.

Go-Blend is currently reliant on a dataset of human playtraces for the environment being used, which is not readily available for most games and is challenging to produce. Reducing the reliance on such data, either partially or completely, through transfer learning or general models of player experience is an important avenue for future work. Furthermore, we currently identify player personas through the game features collected in the AGAIN dataset using simple aggregation methods. A more complex approach could leverage sequential clustering on the playtrace as a whole, or look at alternative representations for clustering such as raw game footage.

Whilst the focus of our experiments was to imitate absolute values for behavior and experience, another promising direction would be to predict changes from one time window to the next (e.g. increase in score/arousal) via preference learning \cite{yannakakis2009preference}.
We would also like to extend this framework to tackle more challenging and stochastic testbeds which are more representative of real-world applications. One approach would be to make use of Go-Explore's robustification phase to train agents capable of performing in stochastic scenarios. In situations where a deterministic setting is not possible for the exploration phase, policy-based Go-Explore could be used to explore the state space and learn a robust policy. Finally, we would also like to investigate the use of more novel methods for generating agent trajectories, such as training a decision transformer \cite{chen2021decision} to efficiently generate human-like trajectories for both behavior and experience.

\section{Conclusions}

This paper introduced the notion of generative AI agents (\emph{procedural personas}) that both play and experience their game as human players do. Our Go-Blend framework leverages Go-Explore's proficiency in hard exploration tasks to reward trajectories for imitating human behavior, imitating human experience, and blending the two. Using the human demonstration data of the AGAIN \cite{AGAIN} dataset, we identify four distinct player personas across 108 players through hierarchical clustering and generate trajectories which imitate their mean score and arousal traces throughout a race. Our results show that Go-Blend is able to identify trajectories which can accurately imitate both behavior and experience simultaneously, as well as exhibit good behavioral patterns even when Go-Blend considers arousal (experience) imitation only. The outcome of this work is a set of trajectories, each expressing distinct play-style and experience patterns that can empower human-like automated play-testing and experience-driven procedural content generation \cite{yannakakis2011experience}.

\begin{acks}
This project has received funding from the European Union’s Horizon 2020 programme under grant agreement No 951911.
\end{acks}

\bibliographystyle{ACM-Reference-Format}
\bibliography{main}

\end{document}